\documentclass[letterpaper, 10 pt, conference]{ieeeconf}  % Comment this line out if you need a4paper

\IEEEoverridecommandlockouts                              % This command is only needed if 
\usepackage{cite}
\usepackage{graphics} % for pdf, bitmapped graphics files
\usepackage{epsfig} % for postscript graphics files
\usepackage{amsmath} % assumes amsmath package installed
\usepackage{amssymb}  % assumes amsmath package installed
\usepackage{amssymb}  % assumes amsmath package installed

\usepackage{xcolor}

\title{\LARGE \bf
Real-Time Dynamics-Based Torque-Sampling MPPI\\for Compliant and Force Aware Manipulation
}
\author{Euncheol Im$^{1,2}$, Taehyun Kim$^{1,2}$, Yonghwan Oh$^{1}$, Myotaeg Lim$^{2}$, Yisoo Lee$^{1,\dagger}$
\thanks{*This work was supported by the National Research Foundation of Korea (NRF) grant funded by the Korea government(MSIT) (RS-2024-00339632), by the Korea Institute of Science and Technology (KIST) Institutional Programs under grant numbers 26E0041, and by Hyundai Motor Company and Kia.}
\thanks{$^{1}$The Center for Humanoid Research, Korea Institute of Science and Technology (KIST), 02792 Seoul, South Korea}
\thanks{$^{2}$The School of Electrical Engineering, Korea University, 02841 Seoul, South Korea}
\thanks{$\dagger$Corresponding author: Yisoo Lee {\tt\small yisoo.lee@kist.re.kr}.}%% <-this % stops a space
}
\begin{document}

\maketitle
\thispagestyle{empty}
\pagestyle{empty}

%%%%%%%%%%%%%%%%%%%%%%%%%%%%%%%%%%%%%%%%%%%%%%%%%%%%%%%%%%%%%%%%%%%%%%%%%%%%%%%%
\begin{abstract}
This study proposes a novel Model Predictive Path Integral (MPPI)-based task-space control framework. The proposed framework explicitly solves rigid-body dynamics within a real-time MPC formulation and enforces safety constraints, enabling accurate motion and force control that yields compliant behaviors for safe and effective physical interaction of robotic manipulators in unstructured environments.
By leveraging MPPI, the proposed framework efficiently handles nonlinear dynamics that are difficult to solve with conventional MPC approaches in real-time.
Furthermore, we develop a torque-sampling-based control architecture that enables efficient exploitation of GPU-based parallelization, resulting in effective compliant and force-aware behaviors.
As a result, the proposed framework achieves a solver update rate of over $166$~Hz with a $0.18$~s prediction horizon, and its performance is validated through real-world experiments on a 7-DoF manipulator.  
\end{abstract}

%%%%%%%%%%%%%%%%%%%%%%%%%%%%%%%%%%%%%%%%%%%%%%%%%%%%%%%%%%%%%%%%%%%%%%%%%%%%%%%%
\section{INTRODUCTION}

% 문제정의 - Dynamics기반의 제어가 중요한데 constraint들을 만족하면서 제어하는게 어려움
As robotic manipulators expand from structured factories to unstructured daily environments, the ability for safe and flexible physical interaction is essential. 
Force control and compliance capabilities are essential for safe physical interaction, particularly in the presence of unexpected collisions and contact-rich tasks.
Dynamics-based control methods, such as inverse dynamics controllers~\cite{khatib2003unified}, have been extensively studied and shown to be effective from a reactive control perspective~\cite{lee2022generalized}.
However, these approaches often lead to locally optimal solutions that may violate physical limits and safety-related constraints.

% Dynamics를 사용하면서 constraint를 고려할 수 있는 MPC -> 단점도 존재
To address this challenge, Model Predictive Control (MPC) has been widely adopted due to its ability to generate optimal trajectories while satisfying complex constraints~\cite{schulman2014motion}.
However, explicitly incorporating full rigid-body dynamics into the optimization loop to ensure dynamic feasibility incurs substantial computational cost, owing to the nonlinear and high-dimensional nature of the system, which poses a major challenge for real-time control~\cite{tassa2012synthesis,neunert2018whole,terry2017comparison,khazoom2024tailoring,norby2024adaptive}.

% MPC 선행 연구들 소개
Quadratic Programming (QP)-based MPC methods have been widely studied for their computational tractability; however, due to the high computational burden, they typically rely on simplifying assumptions such as treating contact forces~\cite{gold2022model} or system dynamics~\cite{terry2017comparison} as constant over the prediction horizon and linearizing inherently nonlinear dynamics~\cite{bednarczyk2020model}.
Despite these approximations, the resulting optimization often remains computationally expensive, leading to short prediction horizons and low update frequencies in practice.
In contrast, Differential Dynamic Programming (DDP)-based MPC methods~\cite{kleff2021high,kleff2022introducing,jordana2024force} aim to reduce model approximations and enable full dynamic control with enhanced computation efficiency; however, these methods remain limited by sensitivity to initialization, convergence to local minima, and the requirement for strictly differentiable cost functions.

To overcome the limitations of optimization- and differential-based MPC methods, sampling-based MPC approaches such as Model Predictive Path Integral~(MPPI) have emerged as a promising alternative for various robotic applications~\cite{williams2017model,williams2018information,seo2023extremely}.
MPPI employs a sampling-based formulation that naturally handles nonlinear dynamics and non-smooth or discontinuous cost functions, while GPU-based parallelization enables highly efficient computation.
However, existing MPPI approaches for robotic manipulation are largely limited to kinematic-level tasks, with sampling typically performed over joint-space variables such as positions~\cite{sundaralingam2023curobo}, velocities~\cite{pezzato2025sampling}, or accelerations~\cite{bhardwaj2022storm,kim2025single}.

Despite the advantages, extending MPPI to explicitly account for full rigid-body dynamics and perform torque-level optimization remains unexplored in manipulation.
Several studies have leveraged numerical simulators, such as Isaac Gym~\cite{makoviychuk2021isaac} for manipulation~\cite{pezzato2025sampling} and MJPC~\cite{howell2022predictive} or Brax~\cite{freeman2021brax} for locomotion~\cite{alvarez2025realtime,xue2024full}, to estimate future system states.
However, these simulator-based approaches may be unable to operate at short control time-steps due to discretization errors~\cite{khazoom2024tailoring} and often remain computationally intensive, which limits their applicability to real-time torque-level optimization in manipulation scenarios.

% Our method
In this study, we propose a novel MPPI-based manipulation framework that enables accurate motion and force control with compliant behavior while satisfying various constraints.
To the best of our knowledge, this work represents the first implementation of MPPI-based manipulation that explicitly incorporates full rigid-body dynamics.
The contributions of this work are as follows.
First, we propose a real-time dynamics-based MPPI framework that directly samples and optimizes joint torques using a tailored GPU-parallelized solver.
Leveraging highly efficient computation, the proposed method achieves an experimentally validated prediction horizon of approximately $0.18$~s with an update frequency of $166$~Hz on a real robotic system, placing it among the fastest dynamics-based MPC approaches capable of both motion and force control reported to date.
Second, we demonstrate through real-world experiments that the proposed framework enables stable and responsive manipulation with dynamics-based capabilities, including hybrid motion-force control, compliance, and feedforward force regulation, under physical interaction scenarios.

\section{Preliminaries: Model Predictive Path Integral}

MPPI~\cite{williams2017model,williams2018information} control is a sampling-based stochastic optimal control method that computes optimal control inputs by minimizing a cost function over a finite time horizon. 
It uses a stochastic system model given by
\begin{equation}
    \mathbf{x}_{t+1} = f(\mathbf{x}_t, \mathbf{v}_t),
\end{equation}
where $\mathbf{x}_t \in \mathbb{R}^{n_x}$ and $\mathbf{v}_t \in \mathbb{R}^{n_u}$ are the state and control input vectors at time $t$, respectively, and $n_x$ and $n_u$ denote the number of states and control inputs.
The control input is represented as
\begin{equation}
    \mathbf{v}_t = \mathbf{u}_t + \delta \mathbf{u}_t, \quad
    \delta \mathbf{u}_t \sim \mathcal{N}(\mathbf{0}, \Sigma_{\mathbf{u}}),
\label{mppi_control_input}
\end{equation}
where $\mathbf{u}_t$ is the nominal control input at time $t$ and $\delta \mathbf{u}_t$ is zero-mean Gaussian noise with covariance $\Sigma_{\mathbf{u}}$, representing disturbances injected into the control input.

The optimal control problem (OCP) aims to find a sequence of control inputs
$\mathbf{u}_{0:T-1}$ that minimizes the total cost over a finite horizon $T$:
\begin{equation}
    \min_{\mathbf{u}_{0:T-1}} S = \phi(\mathbf{x}_T) + \sum_{t=0}^{T-1} \ell(\mathbf{x}_t, \mathbf{v}_t),
\end{equation}
where $\phi(\mathbf{x}_T)$ is the terminal cost and $\ell(\mathbf{x}_t, \mathbf{u}_t)$ is the running cost encoding intermediate objectives such as minimizing control effort, avoiding obstacles, state limitation, or maintaining stability.
In traditional gradient-based optimal control, the above problem is solved by iteratively computing derivatives of the dynamics and cost functions.
In contrast, MPPI solves this OCP in a sampling-based manner, where a total of $K$ noisy control sequences, i.e., rollouts, are evaluated in parallel, and the optimal sequence is updated using an importance-weighted average, as defined by:
\begin{subequations}\label{mppi_optimal_input}
\begin{align}
&\omega_k = \exp(-\tilde{S}_k/ \lambda), \label{mppi_optimal_input:2A}\\
&\mathbf{u}^*_t \leftarrow \mathbf{u}^*_{t,prev} + \frac{\sum_{k=1}^K \omega_k \, \delta \mathbf{u}_{k,t}}{\sum_{k=1}^K \omega_k}, \label{mppi_optimal_input:2B}
\end{align}
\end{subequations}
where $\omega_k$ denotes the importance weight, and $\tilde{S}_k = S_k - S_{\min}$ is the shifted cost of the $k$-th trajectory, introduced to prevent numerical overflow or underflow without affecting the optimality of the algorithm.
Here, $S_k$ is the cost of the $k$-th trajectory, $S_{\min} = \min_{j \in \{1,\dots,K\}} S_j$ is the minimum cost among all trajectories, and $\lambda$ is the temperature parameter.
$\delta \mathbf{u}_{k,t}$ is the sampled noise for the $k$-th trajectory.
After computing the optimal control input sequence, the first control input $\mathbf{u}^*_0$ is applied to the system, while the remaining sequence is used to warm-start the next optimization loop by serving as the nominal control sequence in~\eqref{mppi_control_input}.

\section{Proposed Method}

In this section, we present the proposed torque-sampling MPPI framework. 
The central feature of this framework is the integration of an analytic rigid-body model directly into the optimization process, allowing joint torques to be treated as primary control inputs. 
This torque-level formulation provides a principled foundation for incorporating force-related objectives and supports integrated motion and force control within a single predictive architecture.
\begin{figure}[t]
    \centering
    \includegraphics[width=0.44\textwidth]{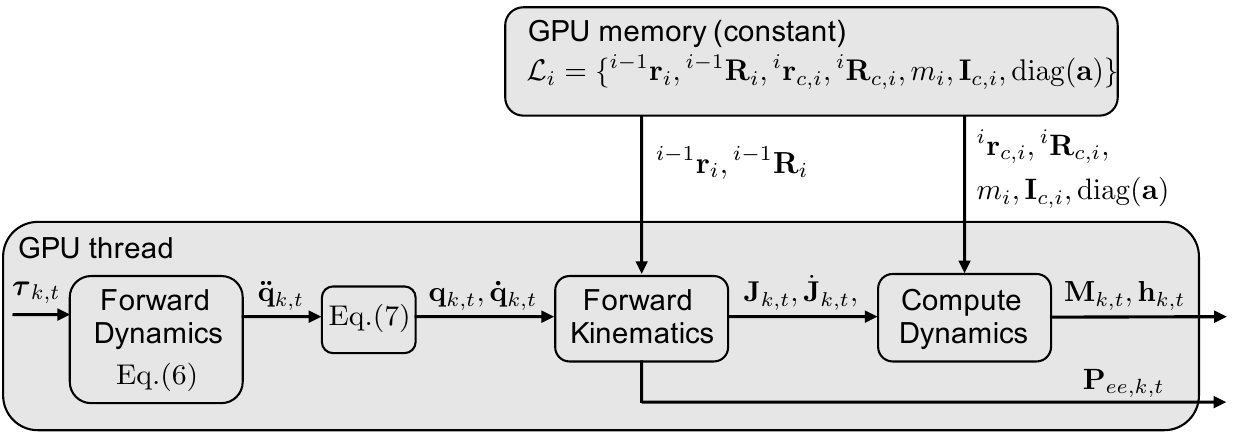} 
    
    \caption{Algorithm flow of the GPU-based dynamics solver. 
    Robot parameters ($\mathcal{L}_i$) are stored in the GPU constant memory, including joint transforms (${}^{i-1}\mathbf{r}_i, {}^{i-1}\mathbf{R}_i$) and Center of Mass (CoM) offsets (${}^{i}\mathbf{r}_{c,i}, {}^{i}\mathbf{R}_{c,i}$), where $\mathbf{r}$ and $\mathbf{R}$ denote the relative position vector and rotation matrix between adjacent frames, respectively.
    Each GPU thread propagates the joint state ($\mathbf{q}_{k,t}$, $\dot{\mathbf{q}}_{k,t}$) and computes the end-effector pose $\mathbf{P}_{ee,k,t}$, the Jacobian matrix $\mathbf{J}_{k,t}$, and the dynamics terms $(\mathbf{M}_{k,t},\mathbf{h}_{k,t})$ at every time step for an independent rollout.}
    \label{fig:structure_CUDA}
\end{figure}

\begin{figure*}[!t] 
\centering
\includegraphics[width=0.85\textwidth]{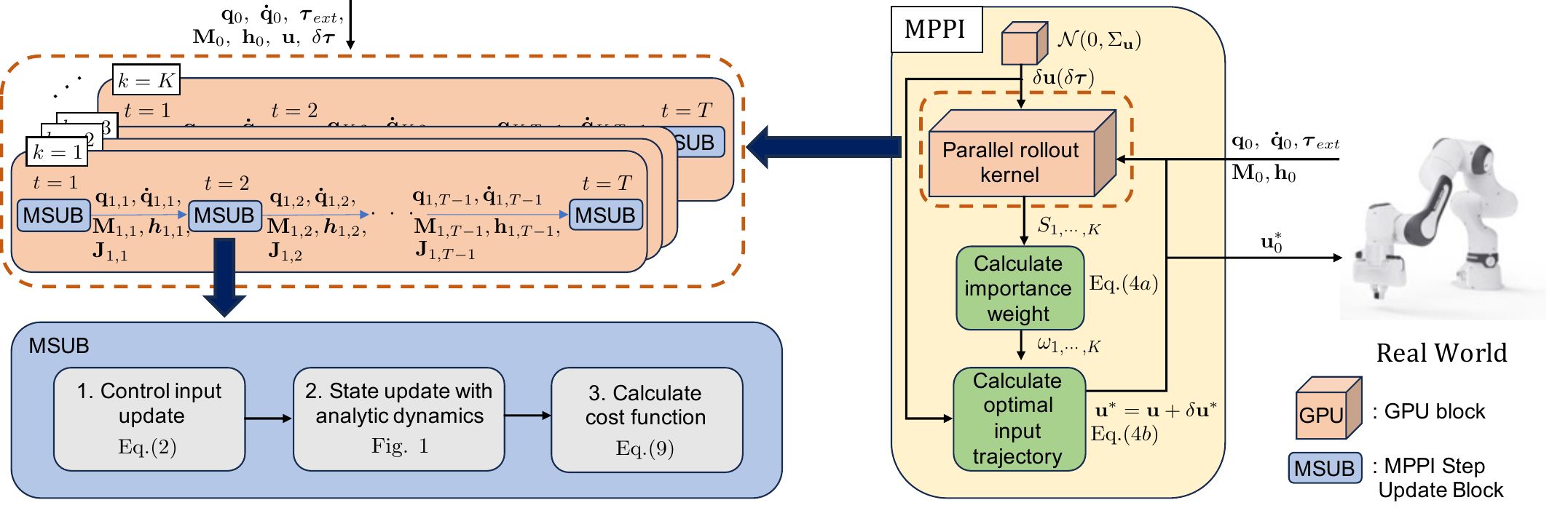} 
\caption{Overview of the control structure of the proposed method, where the orange-colored blocks represent GPU computations and the green-colored blocks represent CPU computations. Specifically, the MPPI Step Update Block (MSUB) computes forward dynamics and cost functions at each timestep for every rollout trajectory to incorporate robot model dynamics.}
\label{fig_structure}
\end{figure*}

\subsection{Analytic Dynamics for Torque-Sampling Rollouts}
The proposed framework employs a torque-level sampling strategy, where control sequences are generated by perturbing the nominal torque: $\mathbf{u}_{k,t} = \boldsymbol{\tau}_t + \delta \boldsymbol{\tau}_{k,t}$. 
Compared to joint acceleration sampling, direct joint torque sampling enables inherent compliance during physical interactions and allows for the seamless integration of force-related objectives within the optimization loop.
To evaluate the cost of each sampled sequence, the framework predicts future states by propagating the robot's configuration over a prediction horizon. 
Given that the control inputs are sampled at the torque level, the framework requires the computation of joint accelerations through forward dynamics before integrating them to update the robot's state.
This relationship is governed by the robot's rigid-body dynamics:
\begin{equation}
    \mathbf{M}(\mathbf{q})\ddot{\mathbf{q}} + \mathbf{h}(\mathbf{q}, \dot{\mathbf{q}}) = \boldsymbol{\tau} + \boldsymbol{\tau}_{ext},
    \label{eq:robot_dynamics}
\end{equation} 
where $\mathbf{M}(\mathbf{q}) \in \mathbb{R}^{n \times n}$ is the inertia matrix and $\mathbf{h}(\mathbf{q}, \dot{\mathbf{q}}) \in \mathbb{R}^{n}$ is the bias force vector comprising Coriolis, centrifugal, and gravitational effects. 
The term $\boldsymbol{\tau}_{ext} = \mathbf{J}^T(\mathbf{q}) \mathbf{F}_{ext} \in \mathbb{R}^n$ represents the projected external joint torque, where $\mathbf{J}(\mathbf{q}) \in \mathbb{R}^{6 \times n}$ is the end-effector Jacobian matrix and $\mathbf{F}_{ext} \in \mathbb{R}^6$ is the external wrench. 
The projected torque $\boldsymbol{\tau}_{ext}$ is assumed to be constant over the prediction horizon.
To update the state for each time step, the joint acceleration $\ddot{\mathbf{q}}_{k,t}$ is explicitly computed by solving \eqref{eq:robot_dynamics} for the sampled torque:
\begin{equation}
    \ddot{\mathbf{q}}_{k,t} = \mathbf{M}^{-1}(\mathbf{q}_{k,t-1}) ( \mathbf{u}_{k,t} + \boldsymbol{\tau}_{ext} - \mathbf{h}(\mathbf{q}_{k,t-1}, \dot{\mathbf{q}}_{k,t-1}) ).
\label{eq:getqacc}
\end{equation}
The joint velocity and position are then updated using an explicit Euler integration scheme with time step $dt$:
\begin{equation}
    \dot{\mathbf{q}}_{k,t} = \dot{\mathbf{q}}_{k,t-1} + \ddot{\mathbf{q}}_{k,t}dt,\qquad\mathbf{q}_{k,t} = \mathbf{q}_{k,t-1} + \dot{\mathbf{q}}_{k,t}dt.
\label{eq:integration}
\end{equation}
For simplicity, $\mathbf{M}(\mathbf{q}_{k,t-1})$ and $\mathbf{h}(\mathbf{q}_{k,t-1}, \dot{\mathbf{q}}_{k,t-1})$ are abbreviated as $\mathbf{M}_{k,t-1}$ and $\mathbf{h}_{k,t-1}$ in the following.

% GPU로 dynamics 계산
To maintain high fidelity throughout the prediction horizon, $\mathbf{M}_{k,t}$ and $\mathbf{h}_{k,t}$ are re-computed at every time step $t$ using the analytic rigid-body model. 
This recursive integration process is implemented directly within custom CUDA kernels to satisfy the high computational throughput required for high-frequency control.
As illustrated in Fig.~\ref{fig:structure_CUDA}, each GPU thread is assigned to an individual rollout to perform forward kinematics; this process determines the link-wise Jacobian matrices and their derivatives, as well as the end-effector pose $\mathbf{P}_{ee}$.
The inertia matrix is constructed by accumulating the transformed link inertias and mass properties:
\begin{equation}
  \mathbf{M}
  = \sum_{i=1}^{n}
    \Bigl(
      \mathbf{J}_{r,i}^{\top}\,
      \bar{\mathbf{I}}_{i}\,
      \mathbf{J}_{r,i}
      \;+\;
      m_{i}\,
      \mathbf{J}_{c,i}^{\top}\,
      \mathbf{J}_{c,i}
    \Bigr)
  \;+\;
  \operatorname{diag}(\mathbf{a}),
\end{equation}
where $\bar{\mathbf{I}}_{i} = \mathbf{R}_{i}\,\mathbf{I}_{i}^{G}\,\mathbf{R}_{i}^{\top}$
is the link inertia tensor rotated into the base frame,
$\mathbf{I}_{i}^{G}$ is the inertia tensor of link $i$ about its center of mass expressed in the link-local frame,
$\mathbf{R}_{i}$ is the cumulative rotation matrix from the base frame to link $i$, $\mathbf{J}_{r,i}$ and $\mathbf{J}_{c,i}$ are the rotational and center-of-mass translational Jacobians of link $i$, and $\operatorname{diag}(\mathbf{a})$ denotes the armature of each joint.
Following a similar accumulation structure, $\mathbf{h}$ is derived from the link-wise Jacobian matrices and their time derivatives.
Since all rollouts share identical robot parameters but differ only in the sampled torque sequences, fixed quantities—such as link masses and kinematic chains—are stored in the GPU's constant memory, enabling parallel computation without inter-thread communication.
This optimized implementation allows for the simultaneous computation of thousands of rollout trajectories, ensuring the real-time feasibility required for high-frequency torque-level MPPI control.

\subsection{Cost Function Design for Hybrid Control}

The proposed torque-sampling MPPI framework explicitly incorporates end-effector force objectives through force-based cost terms, unlike conventional MPPI formulations that focus solely on motion control.
We define $\ell(\mathbf{x}_t,\mathbf{v}_t)$ as a weighted sum of task-related objectives and physical constraints:
\begin{equation}
\ell(\mathbf{x}_t,\mathbf{v}_t) = C_{task} + C_{constraint},
\end{equation}
where the task cost $C_{task}$ is further decomposed into motion and force control components:
\begin{equation}
C_{task} = C_{motion} + C_{force}.
\end{equation}
The motion control cost $C_{motion}$ penalizes deviations from the desired end-effector trajectory in $SE(3)$:
\begin{equation}
    C_{motion} = \mathbf{W}_{pos} \lVert \mathbf{p}_{des} - \mathbf{p}_{ee} \rVert^2 + \mathbf{W}_{ori} \lVert \log(\mathbf{R}_{des}^T \mathbf{R}_{ee}) \rVert^2,
    \label{eq:motion_cost} 
\end{equation}
where $\mathbf{p}_{ee}, \mathbf{R}_{ee}$ denote the current end-effector position and orientation, and $\mathbf{p}_{des}, \mathbf{R}_{des}$ are the desired targets. The log map extracts the rotation vector error.
The force cost is formulated as:
\begin{equation}
C_{force} = \mathbf{W}_{force} \lVert \mathbf{F}_{des} - \mathbf{F}_{ee} \rVert^2 + \mathbf{W}_{reg} \lVert \mathbf{F}_{ee} \rVert^2.
\label{eq:force_cost} 
\end{equation}
In both \eqref{eq:motion_cost} and \eqref{eq:force_cost}, the weighting terms $\mathbf{W}{(\cdot)}$ are diagonal matrices used to specify axis selection and priority. 
To ensure a physically consistent mapping for the end-effector force $\mathbf{F}_{ee} = \bar{\mathbf{J}}^T \boldsymbol{\tau}$, $\bar{\mathbf{J}}^T$ is defined as the transpose of the dynamically consistent generalized inverse~\cite{khatib2003unified}, where $\bar{\mathbf{J}}$ is given by:
\begin{equation}
\bar{\mathbf{J}} = \mathbf{M}^{-1}\mathbf{J}^T(\mathbf{J}\mathbf{M}^{-1}\mathbf{J}^T)^{-1}.
\end{equation}
This formulation accounts for the system inertia, enabling precise task-space force estimation while maintaining decoupling from null-space torques.
Finally, to ensure safety and kinematic feasibility, we incorporate constraint costs defined as:
\begin{equation}
C_{constraint} = C_{joint} + C_{collision}.
\label{cost_constraints}
\end{equation}
The joint cost $C_{joint}$ enforces kinematic feasibility through position and velocity limit penalties, while regularizing the redundant motions of the 7-DoF manipulator to promote balanced postures and prevent convergence to biased local solutions.
The detailed formulations of these costs follow standard practices as described in \cite{kim2025single}.
The collision avoidance cost, $C_{collision}$, accounts for both self-collisions and collisions with external objects.
It is formulated using an indicator function $\mathbb{I}(\cdot)$, which imposes a penalty when the distance falls below the safety threshold:
\begin{equation}
C_{collision} = w_{collision} \sum_{i=1}^L\mathbb{I}(d_i < d_{th,i}),
\label{cost_col}
\end{equation}
where $w_{collision}$ is a scalar collision penalty weight.
The indicator function $\mathbb{I}(\cdot)$ returns 1 if the condition is satisfied and 0 otherwise.
Here, $L$ denotes the number of predefined collision pairs (i.e., link-to-link and link-to-object).
For the $i$-th pair, $d_i$ is the minimum distance computed using the method in~\cite{lumelsky1985fast}, and $d_{th,i}$ is the corresponding safety threshold.

The overall control loop, illustrated in Fig.~\ref{fig_structure}, is executed in parallel on the GPU to ensure real-time performance.
At each time step, $K$ random control noise samples $\delta \boldsymbol{\tau}_{k,t}$ are generated and used to propagate the analytic dynamics according to~\eqref{eq:getqacc} and~\eqref{eq:integration}. Following this parallel rollout, the costs are evaluated, and the optimal control input is updated using the importance sampling rules in~\eqref{mppi_optimal_input}.
The sampling strategy used in this work follows that of~\cite{kim2025single}.

\section{Experiments Results}
%This section validates the proposed torque-sampling MPPI framework through real-world experiments on a 7-DoF Franka Research 3 (FR3) manipulator.
%We evaluate the controller's performance in both free-space and contact scenarios, specifically demonstrating inherent compliance against disturbances, stable contact on curved surfaces, and real-time dynamic obstacle avoidance.

\subsection{Experimental Setup}
Real-world experiments are conducted on a 7-DoF Franka Research 3 (FR3).
The experimental system consists of a GPU-based MPPI controller and a real-time communication interface running on separate computing platforms. A desktop PC (i9-10900KF, RTX 2070 Super) performs the MPPI computation, while an Intel NUC13 (i7, 32~GB RAM) functions as a bridge between the robot and the MPPI PC. The NUC is connected to the robot via the Franka Control Interface (FCI) at $1$~kHz, and it communicates with the MPPI controller via ZeroMQ~\cite{hintjens2013zeromq}. In this setup, the NUC transmits the robot's state to the MPPI PC and relays the updated torque commands back to the robot. The MPPI controller provides updated torque commands at an average rate of approximately $166$~Hz.
The key hyperparameters used in the MPPI controller are summarized in Table~\ref{tab:mppi_params}.
While we verified that the solver can process up to 1,024 rollouts in real time, $K=128$ was empirically selected as it provided sufficient convergence for the given tasks without unnecessary computational redundancy.
To bridge the frequency gap between the solver updates and the $1$~kHz FCI control loop, a zero-order hold (ZOH) was applied to the MPPI torque commands.
\begin{table}[t]
\caption{MPPI Parameters for Different Experimental Conditions}
\label{tab:mppi_params}
\renewcommand{\arraystretch}{1.25}
\centering
\footnotesize
\begin{tabular}{c c c}
\hline
\hline
\textbf{Parameter} & \textbf{Free-space} & \textbf{Contact} \\
\hline
$K$ & $128$ & $128$ \\
$T$ & $30$ & $30$ \\
$dt$ & $0.006$~s & $0.006$~s \\[1mm]

$\mathbf{W}_{pos}$ &
$\mathrm{diag}(0,\,5\!\times\!10^6,\,5\!\times\!10^6)$ &
$\mathrm{diag}(5\!\times\!10^6,\,5\!\times\!10^6,\,0)$ \\

$\mathbf{W}_{ori}$ &
$\mathrm{diag}(5\!\times\!10^5)$ &
$\mathrm{diag}(5\!\times\!10^5)$ \\

$\mathbf{W}_{force}$ &
$\mathrm{diag}(30,\,0,\,0,\,0,\,0,\,0)$ &
$\mathrm{diag}(0,\,0,\,30,\,0,\,0,\,0)$ \\

$\mathbf{W}_{reg}$ &
$\mathrm{diag}(1)$ &
$\mathrm{diag}(1)$ \\

$w_{collision}$ & $30{,}000$ & $30{,}000$ \\
\hline
\hline
\end{tabular}
\end{table}

\subsection{Control Performance Verification}

To evaluate the tracking performance and inherent compliance of the proposed torque-sampling MPPI, we conducted a hybrid motion-force control experiment in free-space, subjecting the robot to external perturbations. The desired task was defined as a step reference: a target force of $F_{x,d} = 10$~N along the x-axis, and target positions of $y_d = 0.0$~m and $z_d = 0.38$~m, while maintaining the initial orientation.
During the operation, several external perturbations were applied to the end-effector to evaluate compliance and force control capabilities.
First, while maintaining a desired force along the $x$-axis, the end-effector motion was intentionally obstructed along the same direction at approximately $t=3\text{--}4.5$~s to assess the response under motion blocking in the force-controlled axis.
Subsequently, external forces were manually applied in the world frame's $+z$, $+y$, and $-y$ directions at approximately
$t=7$~s, $11.5$~s, and $17$~s, respectively.

\begin{figure}[t]
    \centering
    \includegraphics[width=0.44\textwidth]{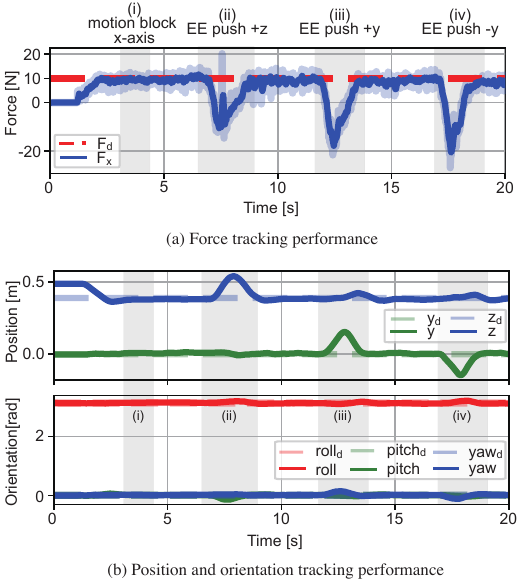} 
    \caption{Experimental results in free space.
    (a) shows the end-effector force tracking, and (b) presents the end-effector position and orientation.
    Shaded regions indicate externally applied disturbances at the end-effector:
    (i) intentional motion blocking along the $x$-axis,
    (ii) external push in the $+z$ direction,
    (iii) external push in the $+y$ direction, and
    (iv) external push in the $-y$ direction, all expressed in the world frame.}
    \label{fig:experiment_freespace}
\end{figure}

The resulting performance is depicted in Fig.~\ref{fig:experiment_freespace}, where the end-effector force is estimated via the torque projection, $\mathbf{F}_{ee} = \bar{\mathbf{J}}^T \boldsymbol{\tau}$, as derived earlier.
Under nominal free-space conditions, the controller successfully tracks the desired force and pose, demonstrating robust performance even under the applied perturbations.
To quantitatively evaluate the tracking performance in the absence of external disturbances,
we computed the average $L_2$-norm errors over the converged interval
($t=4.5\text{--}6$~s), after the removal of motion blocking and prior to disturbance application.
During this interval, the mean $L_2$-norm position error in the position-controlled axes $(y, z)$ was $0.0137$~m, while the mean orientation error was $0.0208$~rad.
When external disturbances are applied, the manipulator responds compliantly in the direction of the applied force
and promptly converges back to the desired force and pose once the disturbance is removed.

\subsection{Hybrid Motion–Force Control under Maintained Contact}

\begin{figure}[t]
    \centering
    \includegraphics[width=0.45\textwidth]{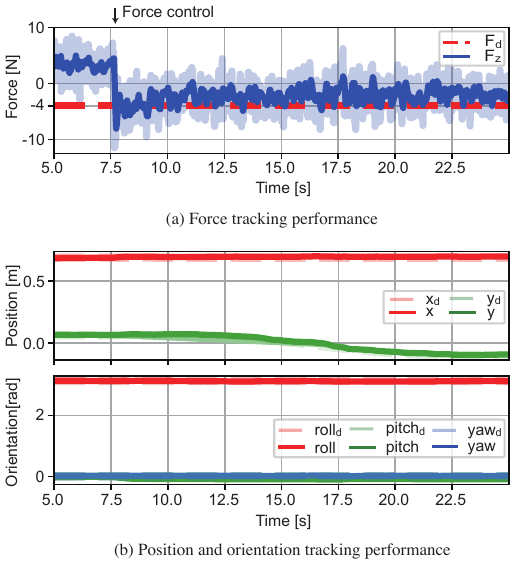} 
    \caption{Experimental results of hybrid motion--force control.
    (a) shows the end-effector force tracking, and (b) presents the end-effector position and orientation during maintained contact.}
    \label{fig:experiment_interaction}
\end{figure}

To evaluate the proposed framework in contact-rich scenarios, we conducted a hybrid motion–force control experiment involving maintained contact with a rigid object.
While the framework was successfully validated on both flat and curved surfaces, we focus here on a cylindrical water bottle, which presents a more challenging scenario due to its curved surface and local geometric irregularities; results for the flat surface experiments are provided in the supplementary material.
To handle this contact, force control was applied along the $z$-axis with $F_{z,d} = -4$~N, while the $y$-axis followed a linear trajectory from $0.06$~m to $-0.1$~m. All other task-space degrees of freedom were regulated to their initial values.
The experimental results in Fig.~\ref{fig:experiment_interaction} are presented starting from $t = 5$~s, including the final portion of the approach phase and the subsequent maintained contact.
During the interval $5 \leq t < 7.5$~s, the end-effector was maintained at a fixed offset above the object using only the task cost $C_{task}$. 
Upon activating the force-related cost $C_{force}$ at $t = 7.5$~s, the controller transitioned to hybrid control, with sustained contact established at approximately $t = 7.67$~s.
Once contact is established, Fig.~\ref{fig:experiment_interaction}(a) demonstrates that the controller reliably maintains the contact state against the curved surface throughout the contact phase.
The force control performance was quantified by a mean absolute error of $1.80$~N over this interval ($t \geq 7.67$~s). 
This error is primarily attributed to the unmodeled curvature of the surface and the absence of explicit force feedback.
Crucially, despite the absence of prior geometric information and dedicated force sensors, our framework achieves stable contact, demonstrating that force-aware behavior emerges directly from the torque-sampling optimization.
While maintaining this contact force, Fig.~\ref{fig:experiment_interaction}(b) also demonstrates stable pose tracking during the interaction phase. 
To quantify this performance, we computed the mean $L_2$-norm errors over the hybrid control interval ($t \geq 7.67$~s), resulting in $0.012$~m in position $(x, y)$ and $0.0921$~rad in orientation. 
These results verify that the proposed framework can effectively manage the trade-off between force and motion objectives in contact-rich manipulation.
\begin{figure}[t]
    \centering
    \includegraphics[width=0.45\textwidth]{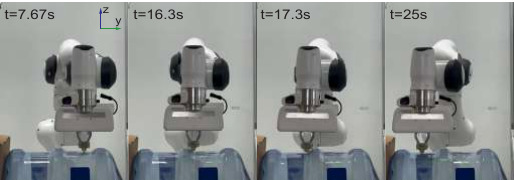} 
    \caption{Sequential snapshots of the hybrid motion--force control experiment on a cylindrical container (water bottle).
    The snapshots illustrate maintained contact and motion along a curved surface at representative time instants.}
    \label{fig:experiment_interaction_snapshot}
\end{figure}

\subsection{Dynamic Obstacle Avoidance}

\begin{figure}[t]
    \centering
    \includegraphics[width=0.45\textwidth]{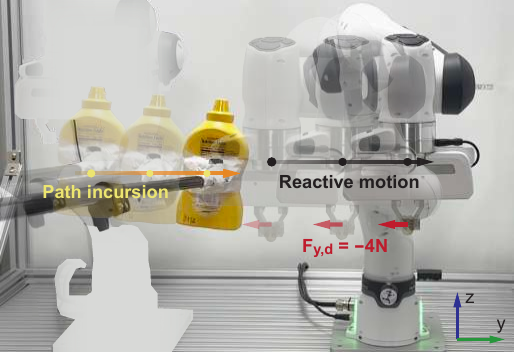} 
    \caption{Experimental results of dynamic obstacle avoidance during hybrid motion-force control. The robot maintains force control on the negative direction of $y$-axis while $C_{collision}$ handles collision avoidance.}
    \label{fig:experiment_obj_avoid}
\end{figure}
The real-time reactivity and safety of the proposed framework were validated through a dynamic obstacle avoidance scenario shown in Fig.~\ref{fig:experiment_obj_avoid}.
The robot was commanded to exert a constant force of $F_{y,d} = -4$~N while maintaining its posture. 
To perceive the environment, an Intel RealSense L515 tracked the obstacle's position at approximately $20$~Hz.
During the task, a dynamic obstacle (a mustard container) was moved in the $+y$-direction, intersecting the robot's path.
As shown in Fig.~\ref{fig:experiment_obj_avoid}, the robot autonomously transitioned its motion toward the $+y$-direction to prevent a collision, momentarily contradicting the primary force command directed toward the $-y$-axis. 
This behavior is governed by the high priority assigned to the collision cost weight $w_{collision}$, ensuring that safety takes precedence over task execution.
For efficient collision detection, the obstacle and robot links were modeled as simplified cylinders with a safety threshold of $d_i = 0.2$~m.
This experiment shows the framework's capability to manage dynamic obstacles while maintaining responsive real-time control.

\section{CONCLUSIONS}
In this paper, we propose a computationally streamlined torque-sampling MPPI framework that integrates rigid-body dynamics directly into the optimization loop.
By evaluating analytic forward dynamics within GPU-parallelized rollouts, the proposed method effectively ensures physical fidelity with a small time-step size while simultaneously achieving a prolonged prediction horizon. 
This breakthrough addresses the fundamental trade-off discussed throughout this work, enabling a high solver update rate of $166$~Hz and a predictive window of $0.18$~s.
% Experimental results on a 7-DoF manipulator validated the framework’s performance in diverse scenarios, including free-space motion under external disturbances and maintained contact tasks.
% The proposed method demonstrates consistent dynamics-based responses, such as compliance and force control, while effectively managing task-space constraints like obstacle avoidance. 
% These results show that by directly incorporating rigid-body dynamics into the torque-sampling process, the framework can handle a wide range of manipulation objectives. 
% This establishes the proposed approach as an effective solution for torque-level control, bridging the gap between high-frequency solver updates and extended prediction horizons without compromising the dynamic fidelity of the system.
Experimental results on a 7-DoF manipulator validate the proposed framework across diverse scenarios, including disturbance rejection in free-space motion and contact-rich tasks.
By directly incorporating rigid-body dynamics into torque sampling, the method enables compliant and force-aware behavior while satisfying task-space constraints, achieving high-frequency updates with extended prediction horizons without sacrificing dynamic fidelity.

In future work, we plan to extend this framework to include object dynamics to enable more sophisticated in-hand manipulation and multi-arm collaborative tasks. 
As these interactions involve complex contact forces, we also aim to address the effects on MPPI caused by dynamic uncertainty and physical contact.

\addtolength{\textheight}{-12cm}   % This command serves to balance the column lengths
                                  % on the last page of the document manually. It shortens
                                  % the textheight of the last page by a suitable amount.
                                  % This command does not take effect until the next page
                                  % so it should come on the page before the last. Make
                                  % sure that you do not shorten the textheight too much.

%%%%%%%%%%%%%%%%%%%%%%%%%%%%%%%%%%%%%%%%%%%%%%%%%%%%%%%%%%%%%%%%%%%%%%%%%%%%%%%%

%%%%%%%%%%%%%%%%%%%%%%%%%%%%%%%%%%%%%%%%%%%%%%%%%%%%%%%%%%%%%%%%%%%%%%%%%%%%%%%%

%%%%%%%%%%%%%%%%%%%%%%%%%%%%%%%%%%%%%%%%%%%%%%%%%%%%%%%%%%%%%%%%%%%%%%%%%%%%%%%%

% \section*{ACKNOWLEDGMENT}

%%%%%%%%%%%%%%%%%%%%%%%%%%%%%%%%%%%%%%%%%%%%%%%%%%%%%%%%%%%%%%%%%%%%%%%%%%%%%%%%
\bibliographystyle{IEEEtran}
\bibliography{MyReference} 

\end{document}